\documentclass{article}

\usepackage{amsmath}
\usepackage{authblk}
\usepackage{hyperref}
\usepackage{amsthm}
\usepackage{latexsym}

\title{Qualitative Decision Methods for Multi-Attribute Decision Making}
\author{Ankit Agrawal} 
\affil{Department of Electrical Engineering and Computer Science \\
McCormick School of Engineering and Applied Science \\
Northwestern University, Evanston, IL, USA \\
\href{mailto:ankitag@eecs.northwestern.edu} {ankitag@eecs.northwestern.edu}  }

\date{}
\begin{document}
\maketitle

\newtheorem{definition}{Definition}
\newtheorem{theorem}{Theorem}

\section{Introduction}

The fundamental problem underlying all multi-criteria decision analysis (MCDA) problems is that of \textit{dominance} between any two alternatives: `Given two alternatives A and B, each described by a set criteria, is A preferred to B with respect to a set of decision maker (DM) preferences over the criteria?'. Depending on the application in which MCDA is performed, the alternatives may represent strategies and policies for business, potential locations for setting up new facilities, designs of buildings, etc. The general objective of MCDA is to enable the DM to order all alternatives in order of the stated preferences, and choose the ones that are best, i.e., optimal with respect to the preferences over the criteria. 

The preferences over the criteria typically represent the relative importance of the various criteria to the DM, which is typically expressed either in the form of an ordering (e.g., a ranking) over the criteria, or if available, a set of numeric weights on each criterion. Most of the existing \textit{quantitative multi-criteria decision methods} (QnMCDMs) \cite{Belton:Springer2002,Figueira:Springer2005} assume that it is possible to (a) completely rank the criteria in order of their importance, and/or (b) precisely quantify the degree of importance of criteria (e.g., as weights). Some formalisms such as Analytic Hierarchy Process (AHP) \cite{Saaty2000,saaty2004decision}, SMART \cite{mustajoki2005decision}, ELECTRE \cite{figueira2005electre,cailloux2010electre,figueira2013overview}, TOPSIS \cite{behzadian2012state,ozturk2011technique}, VIKOR \cite{yazdani2014vikor} and PROMETHEE \cite{behzadian2010promethee,cailloux2010electre,brans2005promethee} elicit relative importance of criteria using a matrix that specifies the pairwise ordering of the criteria, which is then used to generate weights for each criteria by a mathematical transformation \cite{Figueira:Springer2005}. In formal terms, this amounts to assuming that the preferences will always be a \textit{weak order} \cite{Fishburn:Camb1970}. The weak ordering is then typically transformed into a quantitative (numeric) scale, and the transformed importance values are then used to evaluate and compare candidates. 

However, it is not surprising that generation of weights based on qualitative assessments of pairwise relative importance of criteria often leads to instability in rankings of alternatives and their high sensitivity to changing weights or addition of alternatives \cite{Steele:RA2009}. In addition, DMs may not understand the assumptions involved in generating the weights. Moreover, in many applications, the DM's preferences may be naturally incomplete and/or imprecise, especially when the stakes are really high (e.g., trade-offs in facility location for setting up new nuclear reactors). For example, a building designer as a DM may state that increased energy efficiency and pollution efficiency (lesser pollution) are more important than cost efficiency when evaluating candidate designs, without quantifying the relative importance between cost efficiency and either of the other criteria (imprecision), or without even specifying any relative importance between energy efficiency and pollution (incompleteness). A reason for such imprecision may be that precise weights for each criteria may not be available; another reason may be that the DM does not want to commit to precise weights a priori to avoid sensitivity of rankings to initial weight estimation errors. 

\smallskip
In this article, we present and summarize a recently developed MCDA framework that orders the set of alternatives when the relative importance preferences are incomplete, imprecise, or qualitative in nature. To this end, we allow the DM to specify the preference for a decision problem with a set $\mathcal{X} = \{X_1, X_2, ... X_m\}$ of criteria as a \emph{strict partial order} over $\mathcal{X}$, i.e., the relative importance preference $\rhd$ is represented by a binary relation $\rhd \subseteq \mathcal{X} \times \mathcal{X}$. It is easy to see that such a criteria set can be readily represented as a directed acyclic graph. 
We discuss a dominance relation $\succ_d$ recently developed in artificial intelligence literature \cite{Santhanam:JAIR2011}, which has the desirable property that it is a partial order whenever the relative importance preference $\rhd$ is an interval order relation\footnote{For $\rhd$ to be an interval order, it must be irreflexive, transitive, and for all $X_i,X_j,X_k,X_l \in \mathcal{X}, X_i \rhd X_j$ and $X_k \rhd X_l$ must imply that either $X_i \rhd X_l$ or $X_k \rhd X_j$ \cite{fishburn1985interval,myers1999basic}}. 
The ordering of alternatives based on $\succ_d$ is \emph{provably correct}, i.e., if $A \succ_d B$, then it can be argued using principles of rational choice that A is indeed preferred to B. In that sense, the partial ordering produced by $\succ_d$ can be considered as a \emph{reference ordering} that all other rankings must be consistent with. We discuss methods of comparing the orderings produced by $\succ_d$ and other MCDA ranking methods, drawing from recent advances in order theory.

\section{Background}
\label{sec:background}

The following is the typical decision process for multi-criteria decision problems such as the above. 

\begin{enumerate}

	\item First, the stakeholder defines a set of attributes or criteria on which the alternatives can be evaluated and compared. These criteria can be either quantitatively or qualitatively evaluated, e.g., the noise level during construction of a pavement type could be rated on a relative, ordinal scale (Low/Medium/High), or in terms of the maximum decibel level anticipated during construction.

	\item Next, the stakeholder uses his knowledge and experience to evaluate each alternative with respect to each of the attributes on a qualitative scale (e.g., ranking, partial order) or quantitative scale (e.g., utility function), such that the evaluation of two designs with respect to a criterion indicates their relative desirability. Further, the stakeholder may specify relative importance preferences over the criteria quantitatively in terms of numeric compensation (e.g., weights) for the criteria, or qualitatively (e.g., energy consumption is more important than noise levels). 

	\item Finally, an appropriate multi-criteria decision method \cite{EhrgottFigueiraGreco2005} is chosen, which takes into account the evaluations of the alternatives with respect to the criteria, as well as relative importance preference over the criteria and produces an ordering of the alternatives. Each decision method differs primarily in the set of rules of rational choice that it uses to compare alternatives, and hence the orderings they produce may accordingly differ. 

\end{enumerate}

Note that three factors are crucial to the validity and accuracy of the decision: a) the choice of whether each of the criteria will be evaluated on a qualitative or quantitative scale; b) the choice of whether the relative importance preference will be qualitatively or quantitatively specified; and c) the choice of the decision method which determines how the alternatives will be compared with respect to each other based on the stated preferences. 

\subsection{Need for a Qualitative Approach to MCDM}
Quantitative MCDMs, i.e., QnMCDMs such as multi-attribute utility theory (MAUT)\footnote{The notion of utility automatically includes a notion of \textit{uncertainty} which is quantified in the form of probabilities. In this article, we do not consider settings where there is uncertainty over the outcomes of the alternative choices.}~\cite{Keeney:Camb97} require the criteria evaluation scale and the relative importance preferences to be all quantitative. Some other MCDMs such as AHP~\cite{Saaty2000}, SMART, ELECTRE, TOPSIS, and other outranking methods (see~\cite{Figueira:Springer2005}) that focus on certainty allow the stakeholder to specify qualitative input, but eventually convert them into quantitative scores according to some rules of assignment. All the above QnMCDMs apply quantitative decision rules to evaluate dominance, i.e., process (quantitatively elicited or qualitatively elicited and quantified) criteria evaluations and preferences in order to produce a ranking of the alternatives. However, in many design environments there is incompleteness and imprecision in the stakeholder's evaluations and preferences over alternatives. 

\subsection{Limitations of QnMCDMs}
The challenge in using QnMCDMs for decision making in design problems is that when three or more alternatives with multiple attributes are involved, the commonly employed techniques for normalizing evaluations, weighting the criteria and multi-attribute ranking cannot guarantee the selection of the best alternative in an unambiguous, rational, and consistent manner. While QnMCDMs have been largely successful when alternatives can be quantitatively evaluated with respect to criteria and  weights can be precisely estimated, they are inappropriate for problems where some preferences are naturally qualitative as shown below. 

\begin{enumerate}
	\item \textit{Incomplete Preferences: } Stakeholders may assert that two alternatives are incomparable on the same criterion. Similarly, stakeholders may not always be able to totally rank the criteria \cite{Yemshanov:RA2013,Peters:PS2006}. However QnMCDMs assign numeric weights to criteria, assuming that there is a total or weak ordering of the criteria. To deal with incompatibilities, different QnMCDMs extrapolate the stakeholder's preferences and `fill in the blanks' based on certain assumptions to obtain a complete ranking. This questions the validity of use and the accuracy of the results produced by QnMCDMs for the problem.
		
	\item \textit{Semantic Discrepancies: } Numeric weights often have no meaning more than the relative ordering of criteria and the precise meaning of criteria weights is itself often not well understood by stakeholders \cite{Choo:CIE1999,Ross:AIAA2010}. In particular, there is a disconnect between the \textit{intuitive} meaning of the values in the scale as understood by the stakeholder versus the \textit{algorithmic} meaning of the values in the same scale that is assumed by the QnMCDM while comparing alternatives \cite{Belton:Springer2002}. 
	
	\item \textit{Imprecise Preferences: } In some settings, stakeholder may be able to rank alternatives for each criterion and the criteria themselves on their importance, but may not want to commit on the \textit{strength} of such preferences. Thus, evaluation of alternatives on a quantitative scale and weight estimation for criteria in QnMCDMs may involve assumptions extraneous to user input \cite{Yemshanov:RA2013} regarding the extent or strength of preference, questioning the validity of the results.
	
	\item \textit{Complex Tradeoffs: } Stakeholders may tradeoff one set of criteria against another (specify a partial order over \textit{sets of criteria}). For instance, `Improve on \textit{\textbf{both} efficiency and cost}, if possible, even if it increases pollution,' i.e., tradeoff Z against \textit{\textbf{both} X \textbf{and} Y} (but not just one). QnMCDMs are clearly inapplicable in such scenarios, as any weight assignment will violate such a preference.
	
	\item \textit{Possibility of manipulation: } Rankings produced by QnMCDMs are sensitive to changes in the attribute weights, e.g., rank reversals are known to occur even when the attribute weights are kept constant \cite{Steele:RA2009}. This raises the question as to which set of weights is faithful to the preferences of the stakeholder, and makes the results of QnMCDMs vulnerable to manipulation. 
	
	\item \textit{Lack of Transparency: } Stakeholders are often unaware of the mathematical assumptions used in QnMCDMs and their implications on the decisions \cite{Ross:AIAA2010}. The aggregation of criteria based on numeric weights makes it difficult for stakeholders to understand the rationale behind a decision directly in terms of the stated preferences and rules based on principles of rational choice. 
	
	\item \textit{Handling Component Dependencies: } When each alternative is composed of multiple components, the preference over alternatives is a function of the preference over the components that make up the alternatives. Two components may make the alternative highly rated with respect to two different criteria, however when part of the same design, they may cancel out their contributions. For example, a certain choice of wall and paint for a building may be independently preferred, but the particular combination may release toxic gases raising safety concerns. QnMCDMs do not provide a way of accounting for component dependencies in such problems.

\end{enumerate}

\section{Qualitative Preference Reasoning based Decision Method}

We now describe a recently developed qualitative decision making procedure that relies purely on the qualitative relative importance preferences specified by the decision maker and uses mathematical logic to compute dominance and obtain an optimal ordering of candidates. This procedure was first introduced and studied in \cite{Santhanam:JAIR2011,santhanam2010dominance}, and has also been applied to several application domains such as pavement design in civil engineering \cite{Santhanam:ASCE2013}, requirements engineering \cite{oster2011automating}, sustainable design \cite{santhanam2011identifying}, and materials science and engineering \cite{santhanam2014decision}. We now describe how the qualitative decision method computes dominance by formulating and solving logical equations in contrast to existing decision methods, and hence produces an ordering of alternatives based on principles of rational choice.

\subsection{Qualitative Dominance}
\label{sec:qual-dom}
Let $A$ and $B$ be two candidates described by a set $\mathcal{X} = \{X_1, X_2, \ldots X_m\}$ of properties or attributes. Suppose that $\succ_i$ is the intra-variable preference relation on the domain of $X_i \in \mathcal{X}$. For example, if $X_i$ has its domain as the set of all real numbers then $\succ_i$ is often the natural total order on the set of real numbers. Further, let the binary relation $\rhd$ represent the partial trade offs over attributes (relative importance among properties), e.g., Cost $\rhd$ Performance means that Cost is more important than Performance. We say that $A$ dominates $B$ (equivalently, $A$ is preferred to, or is better than $B$), denoted $A \succ_d B$ whenever the following conditions hold: 

\begin{enumerate}
	\item There exists at least one variable, called the witness $X_w \in \mathcal{X}$, with respect to which $A$ is preferred to $B$,  i.e.,  $A(X_w) \succ_w B(X_w)$
	\item $A$ is better than or equal to $B$ with respect to all attributes except those that are less important than $X_w$
\end{enumerate}

If both the above conditions hold, then A is said to be preferred to B with respect to the relative importance tradeoffs of the stakeholder, i.e., $A \succ_d B$. The above conditions can be succinctly expressed as an equation in mathematical logic notation:

\begin{center}
$A \succ_d B \Leftrightarrow \exists X_w \in \mathcal{X} : A(X_w) \succ_w B(X_w) \ \land \ \forall X_k \in \mathcal{X} \ : \ X_w \not\mathrel{\rhd} X_k \Rightarrow A(X_k) \succeq_k B(X_k)$
\end{center}

The semantics or meaning of the above logical equation is as follows: $\Leftrightarrow$ states that the left hand side ($A \succ B$) holds true if and only if the logical formula in the right hand side holds true. The right hand side of the above equation states the existence of a witness property ($\exists X_w \in \mathcal{X} : A(X_w) \succ_w B(X_w)$) such that $A$ is not worse than $B$ for all attributes that are not less important than $X_w$.

The advantage of the above method is that it can handle imprecision in attribute values for alternatives. Hence, it is possible to order alternatives that specify attributes in terms of interval range (as opposed to precise values), or where they are estimated on a qualitative scale (very expensive, expensive, inexpensive, etc.). 

\subsection{Properties of the Dominance Relation}

In order to apply any qualitative decision method to compare and choose among alternatives, it must be checked for conformance to well established principles of rational choice. It is expected that a good dominance relation would satisfy the properties of irreflexivity, transitivity and asymmetry, which makes it a strict partial order. We formally state the definitions of these properties. 

\begin{definition}{Properties of Strict Partial Order Relation \cite{Fishburn:Camb1970}}{}. A dominance relation $\succ$ is:
\begin{enumerate}
	\item irreflexive if for every alternative $A$, $A \succ A$ never holds
	\item transitive if for all triples $A$, $B$ and $C$ of alternatives, if $A \succ B$ and $B \succ C$ both hold, then $A \succ C$
	\item asymmetric if $A \succ B$ then $B \not\mathrel{\succ} A$
\end{enumerate}
\end{definition}

\begin{definition}{Properties of Interval Order Relation \cite{fishburn1985interval,myers1999basic}}{}. A relative importance relation $\rhd \subseteq \mathcal{X} \times \mathcal{X}$  is an interval order if:
\begin{enumerate}
	\item $\rhd$ is a partial order, and
	\item for all $X_i,X_j,X_k,X_l \in \mathcal{X}, X_i \rhd X_j$ and $X_k \rhd X_l$ implies that either $X_i \rhd X_l$ or $X_k \rhd X_j$ 
\end{enumerate}
\end{definition}

It can easily be shown that an irreflexive and transitive relation is always asymmetric, and hence it suffices to show only that $\succ$ is irreflexive and transitive to show that $\succ$ is a strict partial order. We recall the propositions from \cite{Santhanam:JAIR2011}, relating to the properties of $\succ_d$ and state below the main result that qualifies $\succ_d$ as a rational dominance relation. 

\begin{theorem}
 When the input intra-variable preferences $\succ_i$ for each $X_i \in \mathcal{X}$ is strictly partially ordered and the relative importance (trade-offs) over the properties $\rhd$ is interval ordered, the dominance relation $\succ_d$ defined in Section~\ref{sec:qual-dom} is a strict partial order.
\end{theorem}

\subsection{Computational Complexity of Qualitative Decision Making}

The above described qualitative decision procedure involves more computational complexity to compute dominance and to rank the alternatives, in comparison with the 2-objective optimization methods or the MCDA methods like MAUT or AHP. In particular, the computational complexity for evaluating the logical equation in Section~\ref{sec:qual-dom} for each pair of alternatives is of the order of $m^2$ where $m$ is the number of properties or attributes. Hence, if there are $n$ alternatives and $m$ properties from which the best alternative has to be selected with respect to tradeoffs, then it would require the order of $n^{2}m^{2}$ computations to order alternatives, which is expensive (although possible with increased computational power). Despite the complexity, it may be worth using this qualitative method for several reasons. For example, in strategic decisions, one may need precision when incorrect ordering of two alternatives can have a large adverse impact to the decision maker. Another scenario that may need the rigor and precision of qualitative decision method is safety critical decisions such as site selection for harmful chemicals disposal. 

\section{Conclusion}

There are several decision situations where the decision maker needs to order a set of alternatives from the best to the worst, purely based on possibly incomplete, qualitative attributes describing the alternatives, and possibly incomplete qualitative trade-offs among the attributes. We have described a case for using a qualitative decision method in such settings that provides a strict partial order of the alternatives as opposed to linear rankings provided by traditional quantitative decision methods. In the future we will examine the appropriateness of other advances in qualitative decision methods \cite{santhanam2013preference,Santhanam:arxiv2015,Santhanam:JAIR2011,santhanam2010dominance,santhanam2010efficient,santhanam2009dominance} for different applications in engineering \cite{pham2009bensoa,Santhanam:ASCE2013,LKC15,ADC14,MAK14,DGC13,GAC13}, 
optimization-based methods \cite{LKC15,LAL14b,JFW13,JFW15,LAC15}, cybersecurity \cite{santhanam2013verifying,santhanam2013identifying,santhanam2011verifying}, decision-support systems in healthcare \cite{ABR13,BAC13,LTC13,MAF13,ABM13,ARR13,AMN12,AC11a,AC11b,AMN11b}, ranking and recommendation systems \cite{XCZ13c,XCH12b,ZDC13,CAL14,CXC13b,XCH12a}, service and cloud computing \cite{	santhanam2008utilizing,oster2012service,oster2010decomposing,oster2011automating,santhanam2008tcp,santhanam2009web,oster2011identifying,oster2013model}, financial decision making \cite{LAL14a}, and other areas of computing 
\cite{santhanam2015knowledge,ACH11,AMH11,APH13,SCH14}.

\bibliographystyle{plainyr} 
\bibliography{citations} 

\end{document}